# A comparative study of complexity of handwritten Bharati characters with that of major Indian scripts

Manali Naik, V. Srinivasa Chakravarthy

Department of Biotechnology,

IIT Madras, Chennai 600036.

## ABSTRACT

We present *Bharati*, a simple, novel script that can represent the characters of a majority of contemporary Indian scripts. The shapes/motifs of Bharati characters are drawn from some of the simplest characters of existing Indian scripts. Bharati characters are designed such that they strictly reflect the underlying phonetic organization, thereby attributing to the script qualities of simplicity, familiarity, ease of acquisition and use. Thus, employing Bharati script as a common script for a majority of Indian languages can ameliorate several existing communication bottlenecks in India. We perform a complexity analysis of handwritten Bharati script and compare its complexity with that of 9 major Indian scripts. The measures of complexity are derived from a theory of handwritten characters based on Catastrophe theory. Bharati script is shown to be simpler than the 9 major Indian scripts in most measures of complexity.

## INTRODUCTION

India is a land of a large number of languages. There are ten major scripts that are used to write most of the major languages of India. These scripts include: 1) Bengali (used for Bengali and Assamese), 2) Devanagari (used for Hindi and Marathi), 3) Gujarati, 4) Gurumukhi (used for Punjabi), 5) Kannada, 6) Malayalam, 7) Oriya, 8) Tamil, 9) Telugu and 10) Urdu. Most Indian writing systems are based on a peculiar feature known as the composite character or *samyukta akshara* (Daniels and Bright 1996). Unlike linear writing systems like the Roman script used in English, and other Western European languages, where a set of characters are written horizontally, left to right, in a linear fashion, Indian scripts consist of composite characters, which are combinations of smaller units. A single composite character represents either a complete syllable, or the coda of one syllable and the onset of another. [1]

Each of the major Indian scripts listed above (except Urdu) consists of about 16 vowels and about 37 consonants [2] [3]. Tamil has a much smaller number of consonants than other Indian languages.

The Urdu script has an organization that is very different from the remaining 9 scripts. Vowel graphemes display special allographs when they occur in representations of syllables with onsets. These are known as vowel modifiers. Similarly consonant modifiers also do exist.

The problem of communication in India would be immensely facilitated had there been a single language spoken across India. But, in spite of the massive official, nationwide drive to promote use of Hindi, the language is only spoken by about 45% of the current population. A simpler proposition would be seek out a common script, if possible, to write all the major languages currently used.

The possibility of a common script for major Indian languages is meaningful since, 9 of the 10 scripts listed above (Bengali to Telugu) share nearly the same *akshara* structure, barring a few exceptional characters found in individual scripts. Therefore, similar to the situation in Europe, where a common script (Roman script) is used for a majority of European languages, it would be an immense development in the evolution of Indian languages if the entire country can accept a single script. But then a logical and practical approach to choose such a script would be to use one of the 9 existing scripts and add special characters to accommodate the exceptions. However, the question of acceptance of one of the 9 existing scripts by other linguistic communities, to write their own respective languages in that common existing script, is likely to be met with deep social and cultural resistance. Therefore it is a moot point that any one of the existing scripts will be accepted by the entire country. A feasible solution is to develop an altogether new script, a script that possesses advantages not shared by the existing scripts.

In this study, we propose a ‘unified script’ called Bharati which is much simpler and can represent all the 9 major Indian scripts. This study compares handwritten characters of Bharati script to the characters of other Indian scripts by means of measures such as complexity, stability index, stroke density and curvelength.

The outline of the paper is follows. Bharati script is described in Section 2. Section 3 discusses the concepts for evaluation of complexity of handwritten characters of the script. Results of the comparative complexity analysis are described in Section 4.

**The Bharati Script**

A Bharati *akshara* is written in three tiers arranged vertically – a large middle tier flanked by thinner upper and lower tiers. The body of the *akshara* is written in the middle tier. Diacritics that convey vowel modifier information are placed in the upper tier, while diacritics that convey information related to consonants are placed in the lower tier. Both the upper and lower tiers are divided into three regions each, as shown by the dotted lines in Figure 1.

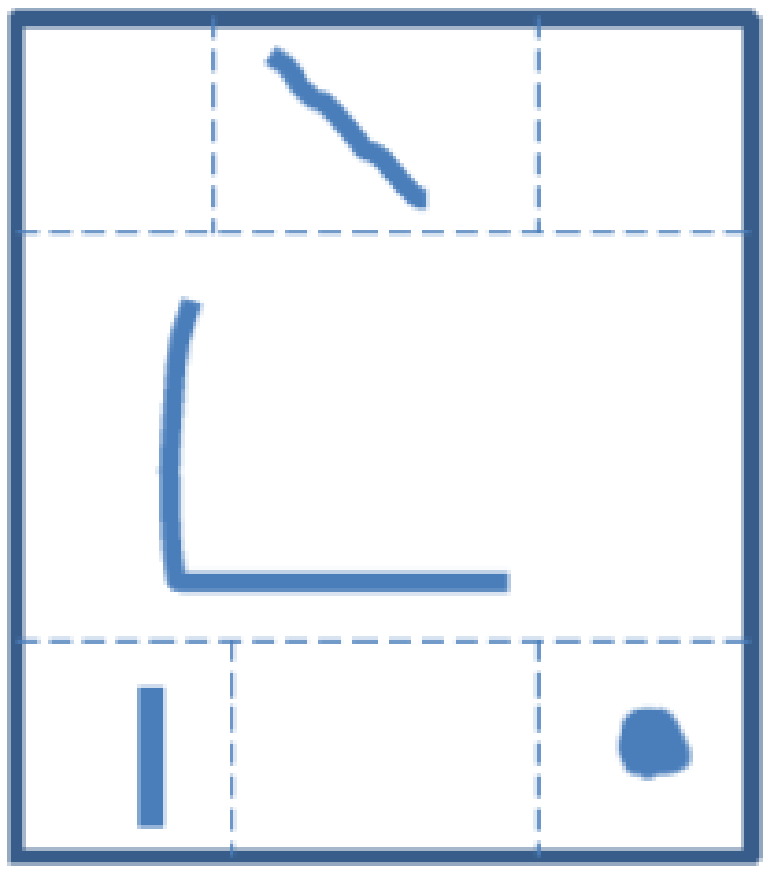

**Figure 1: The three tier structure of a Bharati akshara, Dhe (ढे)**

In its basic organization, Bharati follows the common organization of most Indian alphabet systems into vowels (**अ**, **आ**, **इ**, **ई**…) and consonants (**क**, **ख**, **ग**, **घ**, …). But Indian language alphabets, though ornate, are sometimes unreasonably complicated. In designing Bharati, the underlying phonetic logic of Indian languages is exploited to create a simple script.

To give an example, consider vowels in Devanagari/Hindi script: **अ**, **आ**, **इ**, **ई**,.. Indian language vowels are organized into short (**अ**, **इ**,…) and long (**आ**, **ई**…) forms. Some examples:

1) The long form of **अ** is **आ**; a 'vertical bar' is added to **अ** to produce **आ**.
2) The long form of **इ** is **ई**; a 'hook' is added on top of **इ** to produce **ई**.
3) The long form of **उ** is **ऊ**; a 'hook' is added to the right-middle of **उ** to produce **ऊ**.

Where is the need to have so many different conventions just to denote the long version of a short vowel? There are many such inconsistencies in the design of existing Indian language scripts, which make the characters unreasonably complicated.

Bharati vowels are designed by adding diacritics on top of the vowel 'a'. The diacritics are not arbitrary but follow simple rules that reflect the vowel's phonetic identity. In a sense, Bharati vowels are treated as just another row from the table of Bara Khadi (Consonant-Vowel combinations) characters, wherein the vowel 'a' is some sort of a zeroth consonant.

Long forms of vowels in Bharati script are always constructed by adding a 'horizontal bar' on top of the short form. This rule is followed not only to obtain 'A (आ)' from 'a (अ)', but also for other long forms like 'उ (u)' and 'ऊ (U)', 'ए (e)' and 'एॅ (E)' or 'ओ (o)' and 'ऑ (O)'. The shapes of diacritics are chosen such that their associations in other Indian scripts or even in English/Roman script can be easily identified. For example, the vowel 'उ (u)' is constructed by placing a glyph that resembles 'u' on top of the vowel 'a' (Figure 2). Similarly the vowel 'Ri (ऋ)' is constructed by placing a c-like glyph on top of 'a' which is justified as follows: the vowel-modifier for 'Ri' in Devanagari consists of attaching a c-shaped hook at the bottom of a consonant. To construct the vowel 'ए' we place a diacritic resembling a backstroke on top of the Bharati vowel 'a' (Figure 2). The vowel 'o' (ओ) is constructed by adding a glyph resembling an inverted 'u' on top of 'a'.

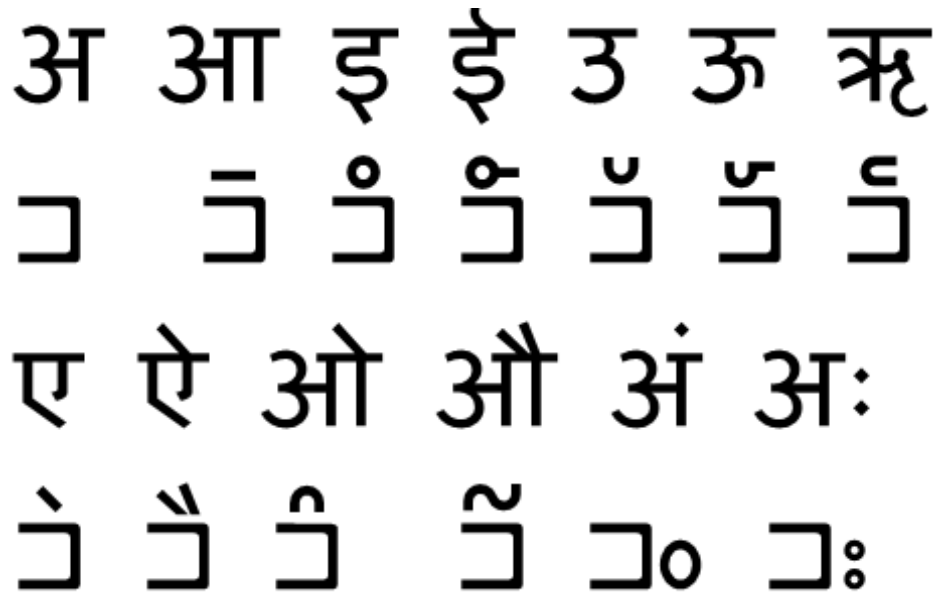

**Figure 2: Devanagari vowels (odd numbered rows) and corresponding Bharati vowels (even numbered rows)**

Construction of Bharati consonants also proceeds on similar lines. Consider the first 25 consonants, organized as a 5 X 5 array (Table 1). The rows are labeled as Velars (V), Palatals (P) etc. denoting the place of articulation of those consonants in the oral cavity. Let us consider an example with the first four velars: 'ka', 'kha', 'ga' and 'gha'. Since the 4 aksharas are variations of the base consonant 'ka' (क), they are represented by placing diacritics under the akshara 'ka'.

Two binary properties (aspiration and voicing) distinguish the first four columns in Table 1. The first akshara in any row is unaspirated and unvoiced, and therefore is taken as a base consonant akshara without any diacritics. We denote 'aspiration' by a dot placed on the right bottom of the base consonant. Similarly 'voicing' is denoted by adding short vertical bar on the left bottom of the base consonant.

| | UA, UV | A, UV | UA, V | A, V | N |
|---|---|---|---|---|---|
| **V** | क k | ख kh | ग g | घ gh | ङ ~N |
| **P** | च ch | छ Ch | ज j | झ jh | ञ ~n |
| **R** | ट T | ठ Th | ड D | ढ Dh | ण N |
| **D** | त t | थ th | द d | ध dh | न N |
| **L** | प p | फ ph | ब b | भ bh | म M |

**Table 1: The first 25 consonants of Devanagari (V = velars; P = palatals; R = retroflex; D = dentals; L = labials; A = aspirated; UA = unaspirated; V = voiced; UV = unvoiced)**

Both the dot and vertical bar are placed to denoted 'aspiration and voicing'. Thus, once the aksharas of the first column in Table 1 are available, the aksharas of columns 2, 3 and 4 may be trivially realized. The fifth column consisting of nasals is handled differently. Nasals are represented by distinct shapes unrelated to the base shape of the aksharas of the first four columns. The nasal of 'ka' family, '~N' (ङ) , is graphically designed as a pruned version of the corresponding akshara in Tamil (ங). To construct the nasal 'NY' (ञ) of the 'cha' (च) family, we begin with ञ, eliminate the prop of the vertical bar, and the shirorekha, and morph whatever is left into a form that can be written in a single stroke without lifting the pen. The nasals 'n' and 'm' resemble the lower case English characters 'n' and 'm' respectively (Figure 3).

The last row of consonants from 'ya' (य) to ' ha' (ह) do not have much redundancy to exploit. There are only two places where such redundancy exists and is exploited as follows.

'l' (ल) and 'L'(ळ) are *liquids*. Therefore 'L' (ळ) is obtained by placing a horizontal bar below 'l' (ल). 's' (स), 'sh' (श) and 'Sh' (ष) are *sibilants*. Therefore 'sh' (श) and 'Sh' (ष) are obtained by placing one and two horizontal bars below 's' respectively.

Once the vowels and consonants are defined, it is straightforward to define CV combinations. The design of vowel modification is identical to how vowels themselves were designed. For example, just as the vowel ‘A (आ)’ is constructed by adding a horizontal bar on the top of ‘a (अ)’, the akshara ‘kaa (का)’ is constructed by adding a horizontal bar on top of ‘ka (क)’. Other aksharas of the Bara Khadi of ‘ka’ are constructed accordingly (Figure 4).

Aksharas of type CVV are not directly supported and are broken up into two aksharas: (C- halant) + (CV). The feature called *halant* in Hindi (or *viraama* in Sanskrit) cancels the inherent vowel (= ‘a’) in a consonant and can be used to break a CCV type akshara (see Figure 5 for examples).

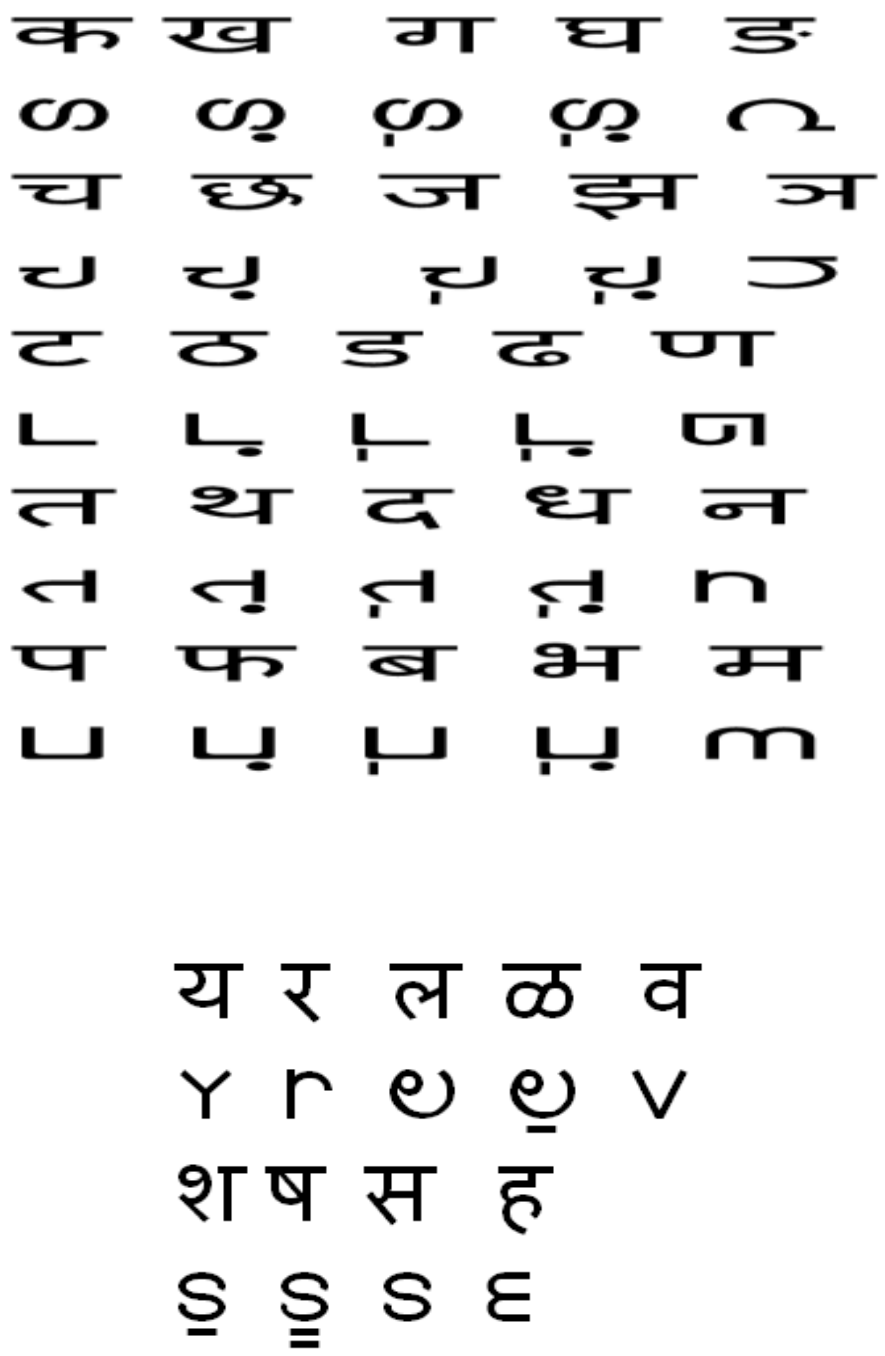

**Figure 3: Devanagari consonants (odd numbered rows) and corresponding Bharati consonants (even numbered rows)**

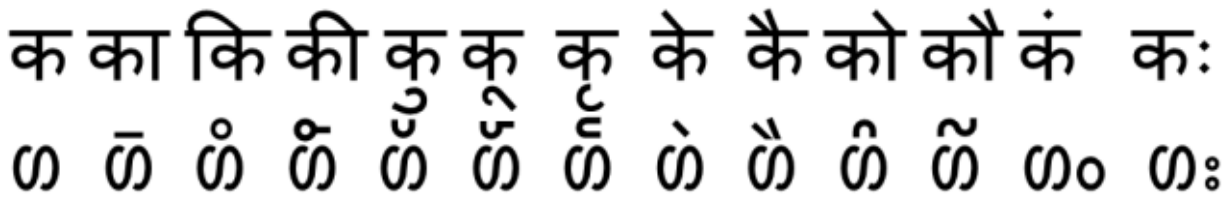

**Figure 4: Devanagari Bara Khadi for 'ka' consonant (odd numbered rows) and corresponding Bharati Barah Khadi (even numbered rows)**

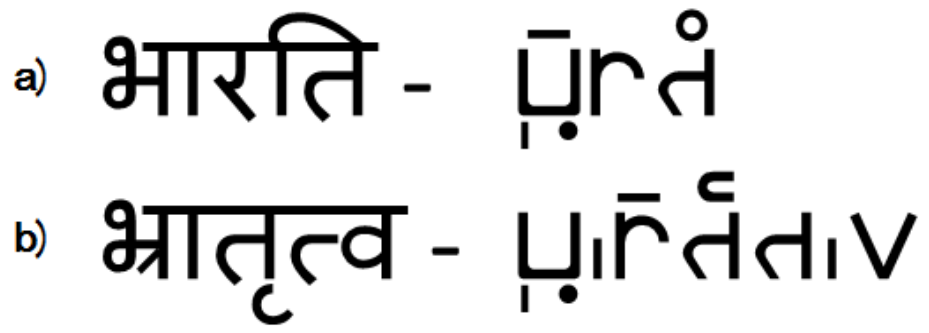

**Figure 5: The words a) bharati (भारति) and b) bhrAtRitva (भ्रातृत्व) written in Bharati script**

### The complexity of handwritten characters

A handwritten character survives serious distortions in size, orientation and even structure. The shape of the character is a feature which survives structural injuries and enables its recognition. We now describe a method of evaluating the complexity of handwritten characters using Catastrophe theory (CT), a branch of Singularity theory. CT aims to formally explain the origin of shapes in Nature[4]-[5], and has been applied to a variety of problems in engineering and physics [4-6]. It investigates and classifies singularities that occur in a special class of dynamical systems called gradient systems, whose dynamics describes gradient descent over a smooth potential function. When such systems are parameterized by a small number of parameters (k<=5), CT shows that the singularities that arise are universal. Furthermore, CT proves that there are only 11 such universal singularities called the *catastrophes*. CT relates such singularities to forms that arise in nature like, e.g. the edge of a breaking wave.

Ideas from CT have been borrowed to represent the shape of handwritten characters (Chakravarthy and Kompella 2003). Since the trajectory of handwriting consists of two functions x(t) and y(t), shape features in the trajectory may be expressed in terms of salient events occurring in x(t) and y(t). According to CT, the overall shape of a smooth function, *f(x)*, is determined by its critical points (CP), the points where the first derivative vanishes. CPs are classified into two categories: (i) simple CPs and (ii) complex CPs. Simple CPs are defined as points where the first derivative vanishes and the second derivative is non-zero. Simple CPs remain the same on small perturbations. Hence, in the neighborhood of simple CPs a function has the property of structural stability[7]. On the contrary, at a complex CP, in addition to a vanishing first derivative, the second and probably other higher derivatives are also zero. In the neighborhood of a complex CP, a function changes its character on a small, smooth perturbation. Hence, it is structurally unstable; on a small perturbation it breaks up into a combination of simple and/or complex CPs [8].

*Codimension* is a parameter that describes the complexity of a complex CP. The codimension of a function near a CP is the minimum number of parameters necessary, in a parametric representation of the function, to bring back the function from a perturbed state to its original state. The higher the codimension of a CP is, the greater the number of parameters necessary to bring back the function to its original state. Therefore, codimension may be regarded as a measure of complexity of a complex CP. This concept can be applied to quantify the complexity of the handwritten characters.

A handwritten character is formed gradually by a sequence of hand strokes [8]. A stroke is defined as what is drawn/written between the time when a pen touches the paper and when it lifts off the paper. Each stroke can be expressed in terms of the x and y coordinates of the trajectory X(t) and Y(t) where t varies from 0 to a maximum time T.

The key idea behind the proposed approach to represent the shape of handwritten characters is that the global shape of a handwritten character may be represented as a graph of a set of local shapes. Furthermore, the local shapes of a handwritten character may seem be classified into a small number of shape classes occurring at points known as Shape Points (SPs).

Below we describe a small number of SPs in terms of X(t) and Y(t) [8]

0) *Interior Point* (I): This is not really a "shape" point but it is important to define it explicitly because it is useful in defining higher order SPs. An *interior point* is simply any interior point of a stroke defined as,

$$X_I = X(t_I),$$
$$Y_I = Y(t_I),$$
$$t_I \in (t_0, t_1)$$

Stability: An interior point is stable since it survives a small, smooth perturbation.

Codimension: Since it is a stable point, its codimension = 0.

1) The End *Point* (E): An end point or a line terminal is the terminating point of a stroke, S, and which does not lie on any other stroke (Figure 6a). An end point in the interval [0, T] is defined as,

$$X_E = X(t_E),$$
$$Y_E = Y(t_E), \qquad (1)$$
$$t_E = \{0, T\}.$$

where no other stroke terminates at $(X_E, Y_E)$.

Stability: An end point is stable.

Codimension: Since it is a stable point, codimension = 0.

2) *Bump Point* (B): A bump point is an interior point where the derivative of either X(t) or Y(t) (with respect to 't') vanishes (Figure 6b). Formally, a Bump point is defined as,

$$X'(t_H) = \left.\frac{dX}{dt}\right|_{t=t_H} = 0; \frac{d^2X}{dt^2} \neq 0; Y'(t_H) \neq 0, or$$
$$Y'(t_H) = \left.\frac{dY}{dt}\right|_{t=t_H} = 0; \frac{d^2Y}{dt^2} \neq 0; X'(t_H) \neq 0. \qquad (2)$$

Near a Bump point, a tangent drawn to a stroke would be either horizontal or vertical. Thus, a Bump point might occur in 4 different ways (Figure 6b).

Stability: The Bump is a simple minimum/maximum of a one dimensional smooth function (X(t) or Y(t)). Being the same as a simple CP of the previous section, it is stable.

Codimension: 0.

The SPs seen so far are the simplest SPs, - all of them are stable. We now define some unstable SPs.

We first define the operation of *identification* useful to describe more complex SPs.

Definition *(Identification)*: Two points A ($X_a$, $Y_a$, $t_a$) and B ($X_b$, $Y_b$, $t_b$) on the same or different strokes, are said to be *identified*, when $X_a = X_b$, $Y_a = Y_b$, $t_a \neq t_b$. Note that if A and B are on different strokes, the constraint $t_a \neq t_b$ is automatically satisfied.

3) *The Cross Point* (X): A cross (X) point can be formed by *identification* of 2 interior points (I) (Figure 6c).

Stability: The 'X' point survives a small perturbation. The actual location of the 'X' point may be displaced, but the point itself remains.

Codimension: Being a stable point, codim = 0.

4) *Cusp Point* (C): The cusp point occurs when both X and Y derivatives vanish simultaneously. In its neighborhood of a cusp point, the stroke has a sharp, spiky appearance (Figure 6d). Formally, it may be defined as,

$$X'(t_0) = \frac{dX}{dt}(t_0) = 0, and, \frac{d^2X}{dt^2} \neq 0, and$$
$$Y'(t_0) = \frac{dY}{dt}(t_0) = 0, and, \frac{d^2Y}{dt^2} \neq 0. \quad (3)$$

Stability: At a Cusp point, derivatives of both X and Y functions vanish simultaneously, i.e., $X'(t_c) = Y'(t_c) = 0$. On a small perturbation to X(t) and Y(t), it may so happen that, $X'(t_a) = Y'(t_b) = 0$, where $t_a$ and $t_b$ are unequal. Thus, where there was a Cusp earlier, we now have 2 Bump points (one along X and another along Y). Hence the Cusp point is unstable. On perturbation, a Cusp may change into a smooth bump or a self-intersecting loop (Figure 6d).

Codimension: Since we only need a single degree of freedom – vary either $t_a$ or $t_b$ until it equals the other - to make the 2 Bump points coincide, codimension of a Cusp = 1.

5) *The "T" point* (T): A T point is formed by identification of an interior point (I) and a terminal point (E) (Figure 6e).

Stability: It is obvious from fig. 6e that 'T' is unstable.

Codimension: The "T" can be restored by moving the end point, in Fig. 6, in only a single dimension. Therefore codimension = 1.

6) *Dot Point* (D): The Dot SP arises from the simplest kind of stroke – a stroke of zero length (Figure 6f), given as:

$$X = X(t); \quad Y = Y(t); \quad t \in [t_0, t_1], \text{ where } t_0 = t_1. \quad (4)$$

Stability: This is an unstable point since on a small perturbation the zero-length stroke may turn into one of non-zero length.

Codimension: To turn the non-zero length stroke back into a Dot, only one parameter ($t_0$ or $t_1$) need to be modified, so as to make the inequality between $t_0$ and $t_1$ into an equality. Hence codimension = 1.

7) *The Angle Point* (A): An angle point is formed by identification of two End points (E) (Figure 6g). It occurs when a stroke begins from where another stroke had ended.

Stability: From the perturbed forms shown in Fig. 6g, it is obvious that the Angle point is unstable.

Codimension: The Angle can be restored by moving one of the End points, in Fig. 6g *with two degrees of freedom*. Therefore codim = 2.

Among the 7 SPs introduced so far, there are 3 SPs (E, B and X) with codim = 0, and 3 SPs (D, C, T) with codim = 1. In a more complete description of SPs given in (Chakravarthy and Kompella 2003), the number of SPs with codim = 2 is greater than 6, but here we present only 1 of them. Of all the SPs with codim = 2, we found that only the Angle occurs in Indian language characters and Bharati characters.

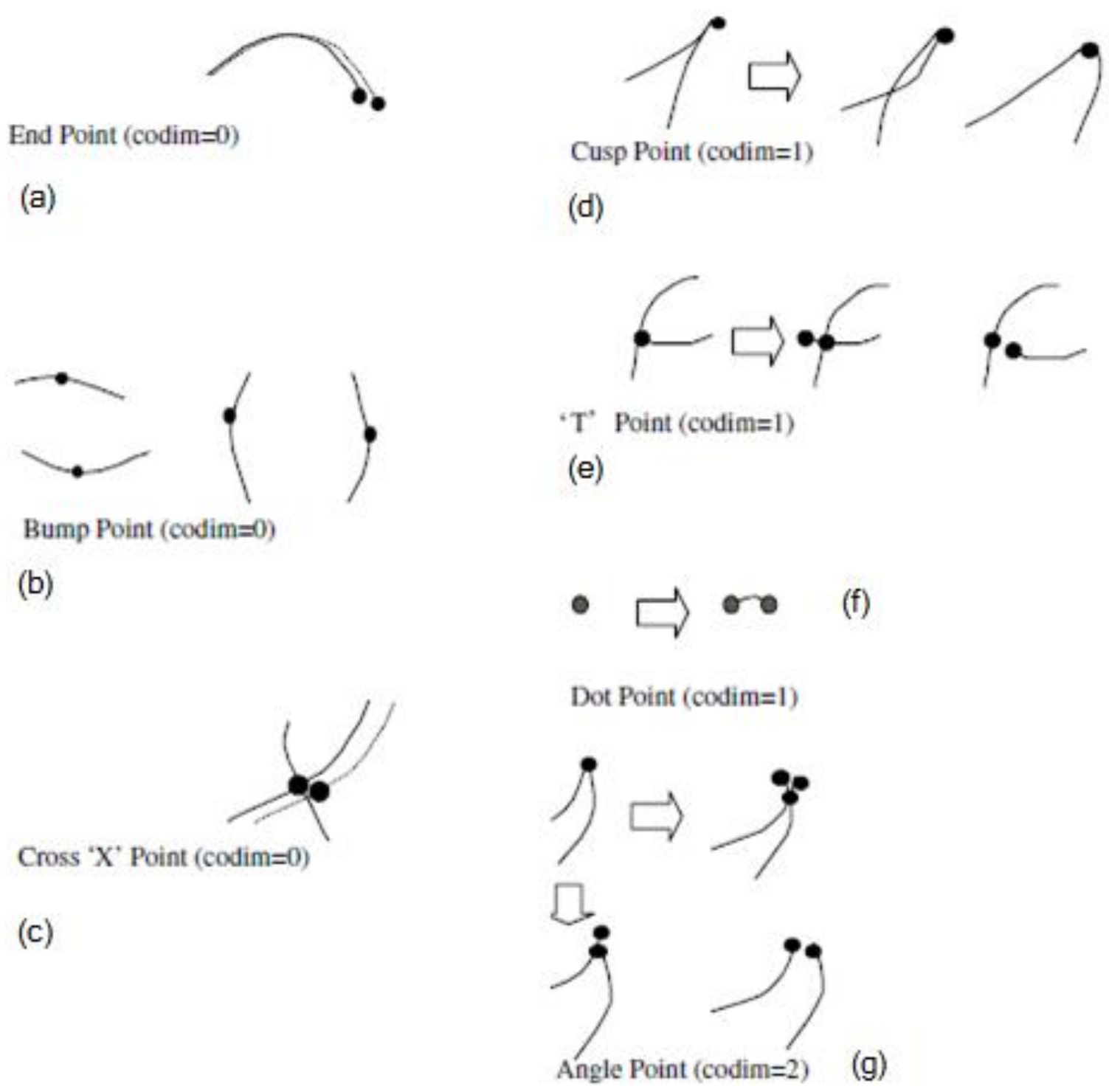

**Figure 6: Illustrations of stable (LT, B and X) and unstable (C, T, D and A) shape points**

The Critical points have been defined for handwritten strokes, which are referred to as shape points (SP) hereafter. SPs with codimension value equal to zero are stable shape points. Line End point (E), Bump point (B) and Cross point (X) are considered here. Cusp point (C), T-point (T), Dot point (D) and Angle point (A) has non-zero codimension value. These shape points are structurally unstable shape points [8]. Next we describe a method of assessing the complexity of handwritten characters and apply the same to handwritten text written in the 9 Indian language scripts of interest, and compare the complexity results with the corresponding results from Bharati.

## METHODS

Along with Bharati, the following nine different Indian scripts are considered for this study:

1) Bengali, 2) Gujarati, 3) Hindi (Devanagari script), 4) Kannada, 5) Malayalam,

6) Oriya, 7) Punjabi (Gurumukhi script), 8) Tamil and 9) Telugu.

Twenty names of Indian cities covering all the vowels and consonants were selected carefully (see Table 2: List of names of cities (in English and Devanagari script) used for data collection). The names of cities were written in each script by the writers using a digital pen.

*Hi-Tech e-Writemate* digital pen was used to capture and store handwritten data. The data obtained using digital pen represents the x- and y - coordinates of the strokes of the handwritten characters. The following preprocessing steps were applied to the data obtained:

(1) Character segmentation: The strokes are segregated according to their horizontal and vertical position and stored as a structure to represent a single handwritten character. Figure 7: Character segmentation for Hindi (Devanagari script); strokes with same colour in each row present a single character shows segmented characters of Hindi (Devanagari script) represented with different colours.

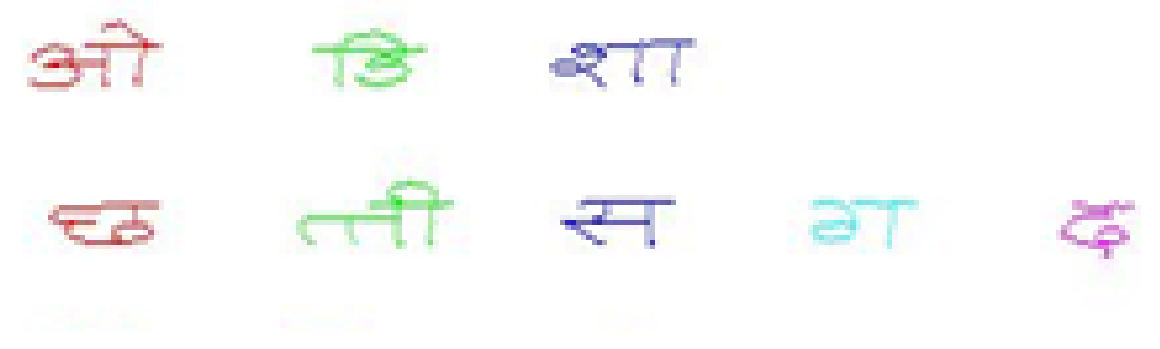

**Figure 7: Character segmentation for Hindi (Devanagari script); strokes with same colour in each row present a single character.**

(2) Normalization: The characters in most of the Indian scripts are written using more than one stroke. Hence, the characters written in these scripts are of different height and width. The characters are normalized in size by scaling the x- and y-coordinates of the strokes using the same factor. The factor considered for normalization is the height of the main stroke. The main stroke of a character is the stroke with the largest y-span [9].

A representative example of handwritten character with strokes normalized based on the height of the main stroke is shown in Figure 8: Normalization of strokes of Hindi handwritten character, au (औ) , based on the height of the tallest stroke

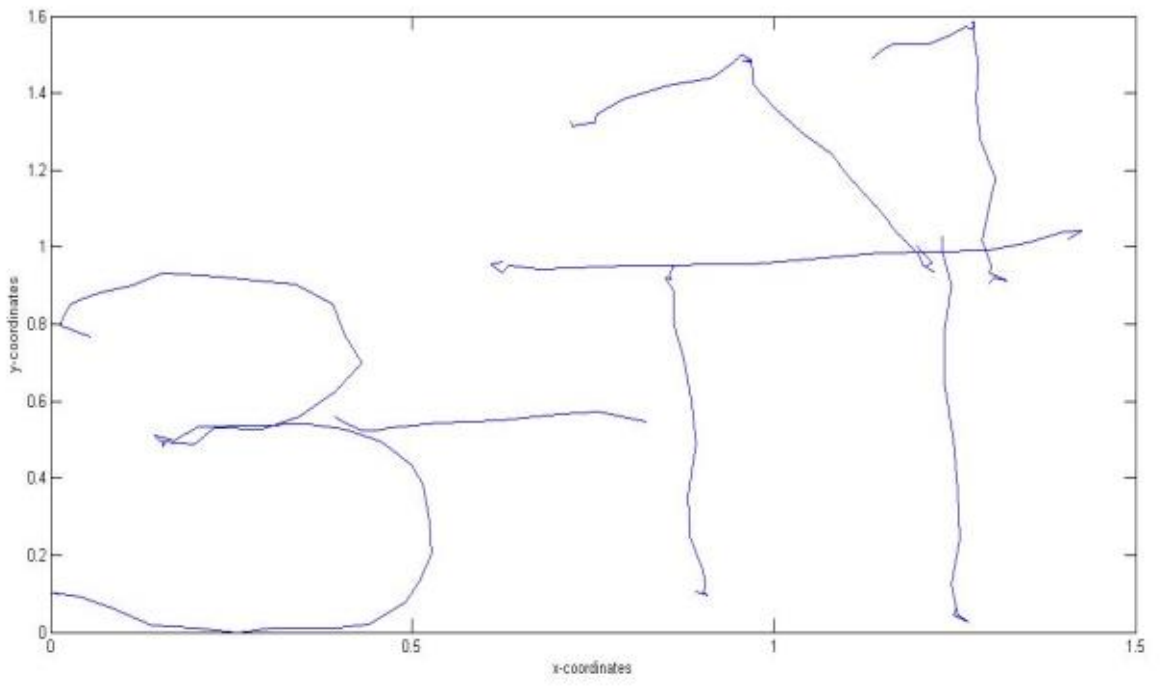

**Figure 8: Normalization of strokes of Hindi handwritten character, au (औ) , based on the height of the tallest stroke**

(3) Smoothing: SPs defined in the previous section involve derivative computations, which require X(t) and Y(t) to be smooth. Smoothing of strokes is achieved by convolving X(t) and Y(t) with a one dimensional Gaussian kernel, g(u) defined below:

$$g(u) = \frac{1}{\sqrt{2\pi}\sigma_s} e^{-\frac{(u-\mu)^2}{2\sigma_s^2}}, 1 \le u \le 21 \quad \textbf{(5)}$$

where $\mu$ is the center of the Gaussian function, and $\sigma_s$ is the width parameter of the Gaussian function [9]. The effect of smoothing can be seen in Figure 9.

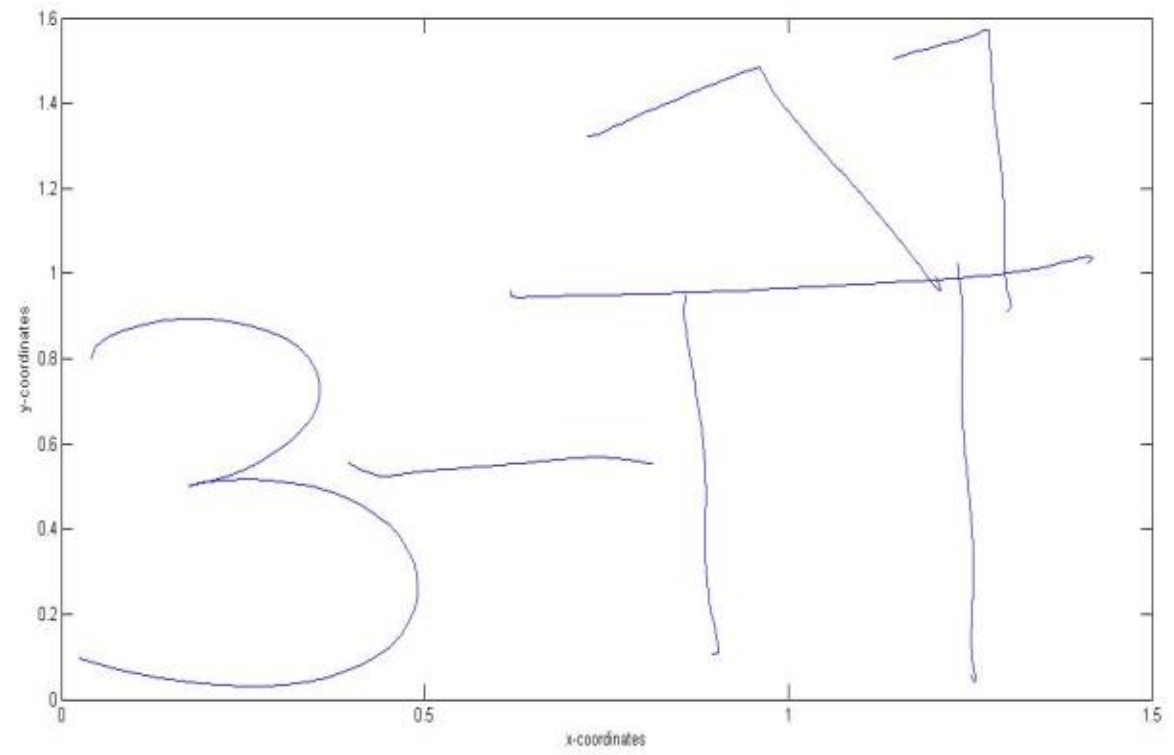

**Figure 9: Smoothing effect on a Hindi handwritten character, au (औ)**

(4) Interpolation: The final step in preprocessing is where the smoothened stroke is interpolated to give a fixed number of points, equally spaced along the curve length. The number of points is chosen based on the average number of points per stroke in the given dataset. A linear method of interpolation was used to get 64 equally spaced points along the curve. The interpolated strokes, representing the handwritten characters, were used for the identification of the SPs. MATLAB functions were written to identify stable and unstable SPs on the strokes. Figure 10 shows an example of a character labeled with End points or line terminal points (l1 and l5), bump points (b1, b3 and b7), cusp point (c) and cross point (X).

| | |
|---|---|
| Rajasthan (राजस्थान) | Odisha (ओडिशा) |
| Aurangabad (औरंगाबाद) | Chhattisgarh (छत्तीसगढ) |
| Udaipur (उदयपूर) | Bharuch (भरूच) |
| Sindhudurg (सिंधुदुर्ग) | Thane (ठाणे) |
| Meghalaya (मेघालय) | Amritsar (अमृतसर) |
| Ernakulam (एर्नाकुलम) | Aagra (आग्रा) |
| Aizawl (ऐजोल) | Itanagar (इटानगर) |
| Ambarnath (अंबरनाथ) | Mumbai (मुंबई) |
| Jharkhand (झारखंड) | Umbergaon (ऊम्बरगाव) |
| Sriharikota (श्रीहरीकोटा) | Fatehpur Sikri (फतेहपुर सिक्री) |

**Table 2: List of names of cities (in English and Devanagari script) used for data collection**

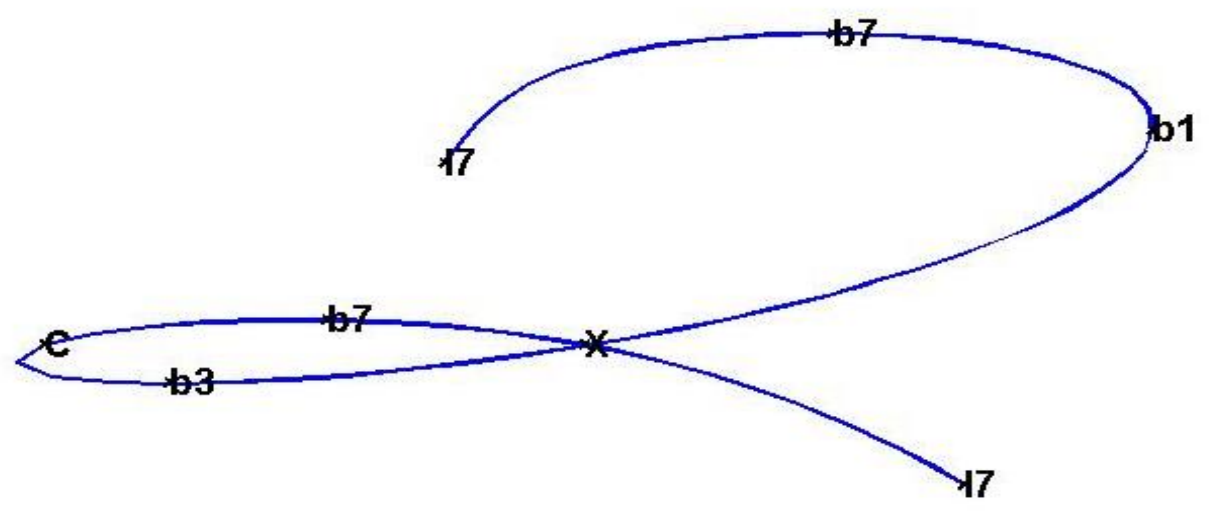

**Figure 10: Gujarati character (r) labeled with shape points; stable shape points- line terminal point (l7), bump point(b1, b3 and b7) and cross point (X), unstable shape point – cusp point (C)**

We now define a set of measures of complexity using which we define the 9 Indian scripts with Bharati.

<u>Shape Complexity:</u>

We now define the Shape Complexity of a stroke as the sum of complexities of all the SPs. The complexity of a SP is defined as,

$$\text{Shape Complexity} = 1 + \text{codimension}, \qquad (6)$$

where codimension is the codimension of the SP. A stroke with more SPs (even if all of them are structurally stable) is considered more complex than one with fewer SPs. Therefore the shape complexity of an SP with codimension = 0 is defined as 1. Hence, the net shape complexity of a character is calculated as,

$$\text{Net Shape Complexity} = \sum_{i} (1 + \text{codimension}_i) * N_i \qquad (7)$$

where, i represents the type of SPs ( i = E, B, X, C, T, A or D), $\text{codimension}_i$ represents the codimension value for shape point i and $N_i$ represents the total number of shape points of type i identified in the handwritten characters of all the 20 city names. Since E, B, X, C, T, A, and D are most commonly occurring SPs in handwritten characters, we consider only these 7 SPs in the present study.

Shape Complexity (as per equation Net Shape Complexity $= \sum_i (1 + \text{codimension}_i) * N_i$ **(7)**

depicts the Net Shape Complexity for the entire set of 20 words written in various scripts. But it is desirable to calculate Shape Complexity per Unicode which denotes the complexity density of the characters in the script of interest. The number of unicodes in the word gives us information about the total number of vowels and consonants in the word. The total Spatial Complexity of all the words written in a given script is divided by the number of unicodes to yield "Shape Complexity" which is estimated for all the 10 scripts.

Thus we define,

Shape Complexity = Net Shape Complexity/#Unicodes (8)

Furthermore, the actual value of Shape Complexity (eqn. Shape Complexity = Net Shape Complexity/#Unicodes (**8**) depends on the SPs considered in the calculation. Based on the selection of SPs used in calculation of Shape Complexity, we define 3 complexity measures:

Shape Complexity #1 = Shape Complexity calculated using all of the 7 SPs considered.

Shape Complexity #2 = Shape Complexity calculated only using the 6 of the 7 SPs (excluding E) considered.

Shape Complexity #3 = Shape Complexity calculated only using only the 3 unstable SPs (C, T and D) of the 7 SPs considered.

<u>Curvelength:</u>

The total curvelength of a word in its size-normalised form is a reasonable measure of complexity of the word. Curvelength of the character is the sum of the curve-lengths of all the strokes in a handwritten character [9]. Hence, it can be used as one of the measures for the comparison across all the scripts. The interpolated strokes were used for calculation of the curvelength (as per equation Curvelength **per** character =

$$\sum_j \sum_{i=1}^{i=N-1} \sqrt{(x_{i+1} - x_i)^2 + (y_{i+1} - y_i)^2} \quad (\mathbf{9}).$$

Curvelength per character =

$$\sum_j \sum_{i=1}^{i=N-1} \sqrt{(x_{i+1} - x_i)^2 + (y_{i+1} - y_i)^2} \quad (9)$$

where, N represents the total number of points in an interpolated stroke, *j* represents the total number of strokes in a handwritten character. Thus, the curvelength of a character is the sum of the curvelengths of all the strokes in the character.

Stability Index

We define the stability of a character in terms of the number of structurally stable SPs (codimension = 0) that the character has relative to the number of structurally unstable SPs (codimension > 0). Scripts whose characters possess more structurally stable SPs are likely to be more stable. Among the 7 SPs considered in the present study (E, B, X, C, T, A, and D), E, B and X are structurally stable, while the rest are unstable. Therefore, Stability Index is defined as.

$$\text{Stability index} = \frac{\textbf{Total number of stable shape points}}{\textbf{Total number of unstable shape points}} = \frac{\mathbf{E+B+X}}{\mathbf{C+T+D+A}} \quad (10)$$

The above calculation is performed over the entire set of 20 words for each script.

## RESULTS

The stable and unstable SPs for handwritten characters in ten scripts were identified. The total number of strokes required for writing twenty names of cities is lower for South Indian scripts like Tamil, Malayalam and Telugu. This number is higher for North Indian scripts like Hindi and Punjabi. Total number of strokes for Bharati script falls in the average range. The four South Indian scripts – Kannada, Malayalam, Tamil and Telugu, - have the lowest scores in this respect, which reflects the popular understanding that South Scripts are ornate, with complex, convoluted strokes. Among South Indian scripts, single strokes often represent and entire CV combination, which explains the low value of strokes/Unicode for these scripts (**Error! Reference source not found.**).

However, the results are different for curvelength measure. Bharati script scores the lowest among all the scripts for measures as curvelength, curvelength per unicode (**Error! Reference source not found.**). Gujarati emerges as a runner up after Bharati in this measure.

Stability Index, defined as (E+B+X)/(C+T+D+A) (eqn. **Stability** index = $\frac{\textbf{Total number of stable shape points}}{\textbf{Total number of unstable shape points}} = \frac{\mathbf{E+B+X}}{\mathbf{C+T+D+A}}$ (**10**), is compared across all the 10 scripts. Stability index (**Error! Reference source not found.**) was found to be highest for Bharati with Telugu in the second position.

Three complexity measures – Complexity #1, Complexity #2, and Complexity #3, - were computed for all the 10 scripts. Bharati script was found to have the smallest value for all the 3 measures. (**Error! Reference source not found.**, **Error! Reference source not found.**, **Error! Reference source not found.**).

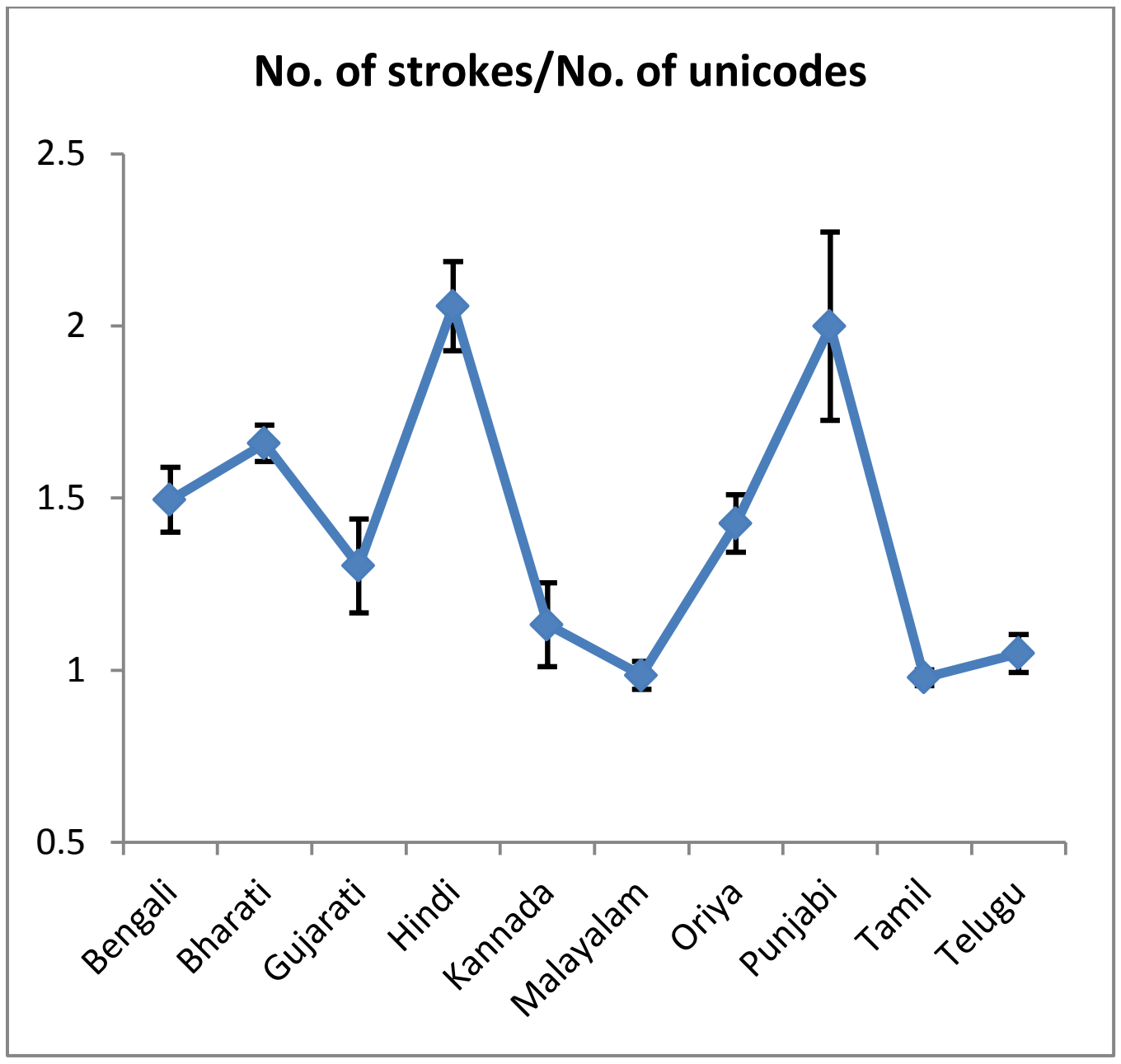

**Figure 11: Total number of strokes per unicode; Bharati script falls in the middle range**

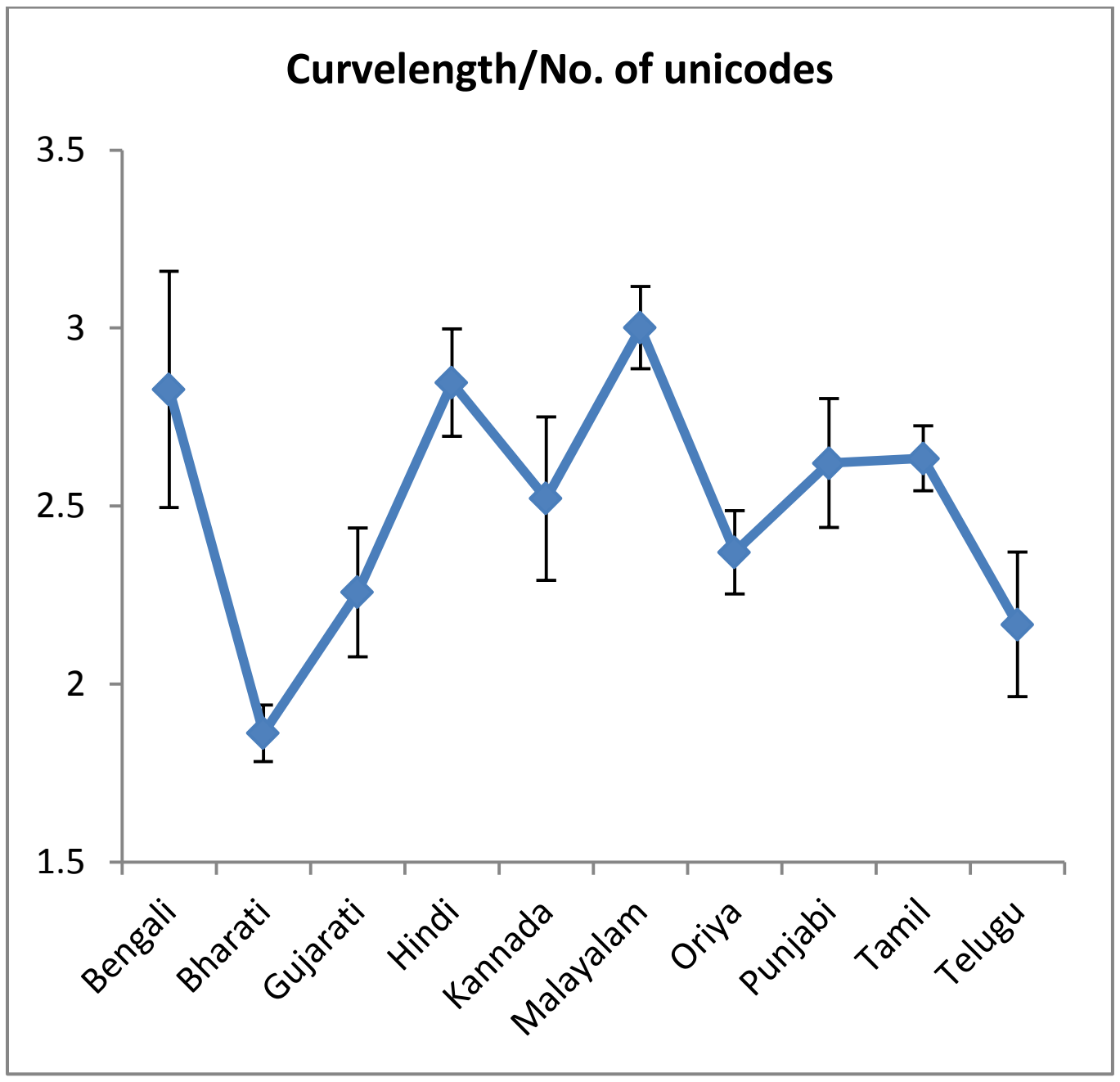

**Figure 12: Curvelength per unicode for characters of ten scripts; lowest value for Bharati script**

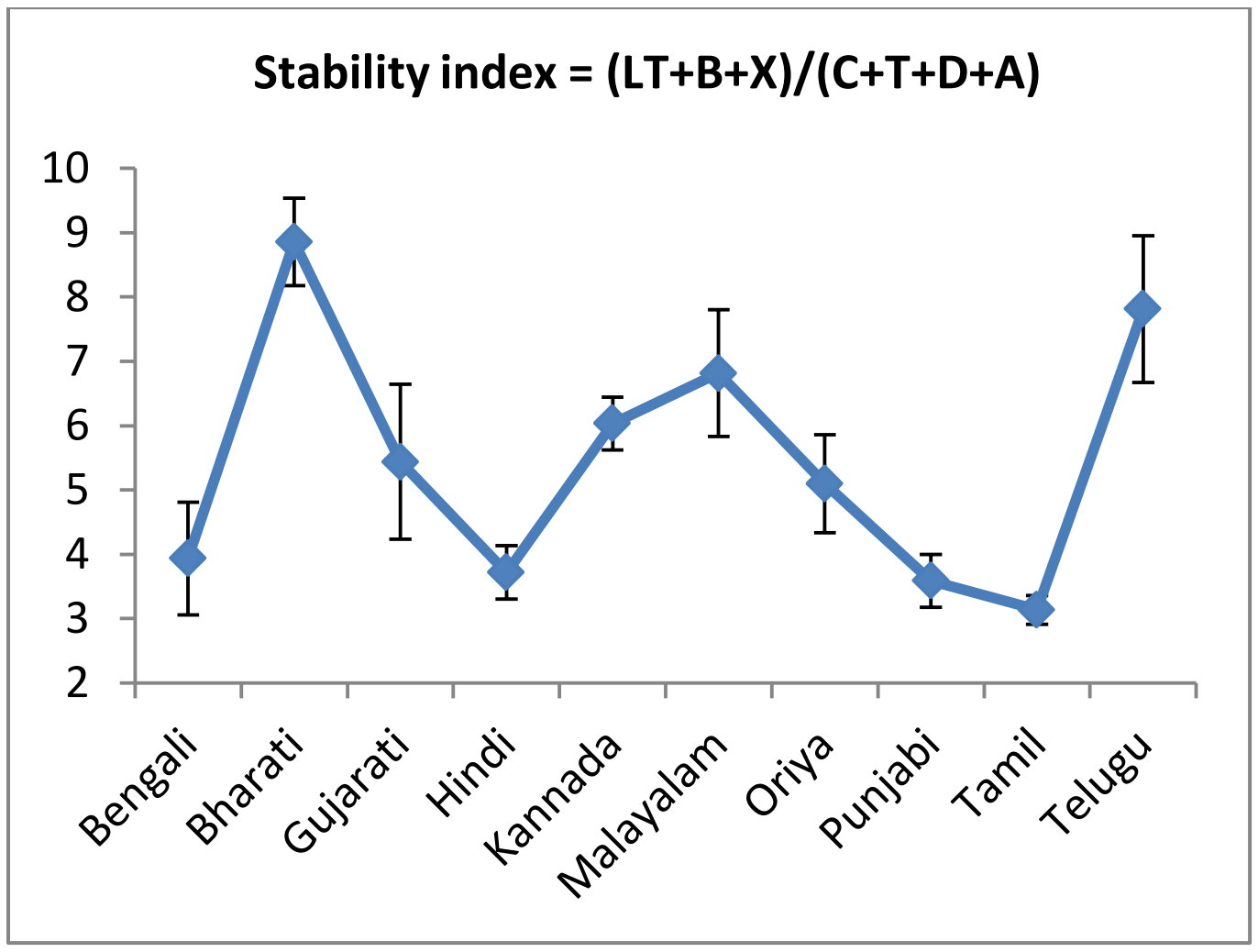

**Figure 13: Stability index, a ratio of total number of stable shape points and total number of unstable shape points; highest value for Bharati script; lowest value for scripts like Hindi and Tamil (rich in T and C points)**

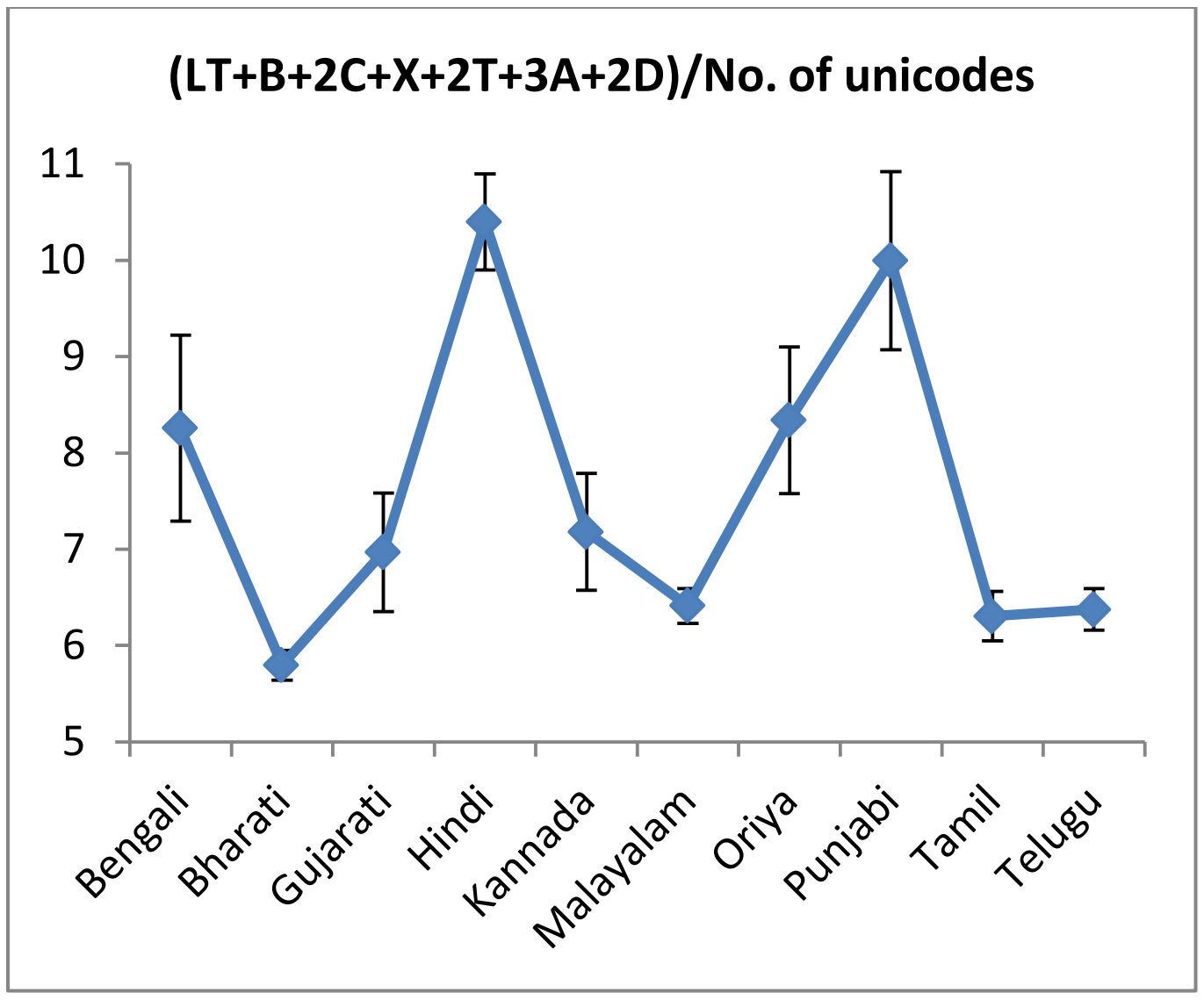

**Figure 14: Complexity #1 is based on 7 SPs (E, B, C, X, T, A, D). The least value is observed for Bharati script**

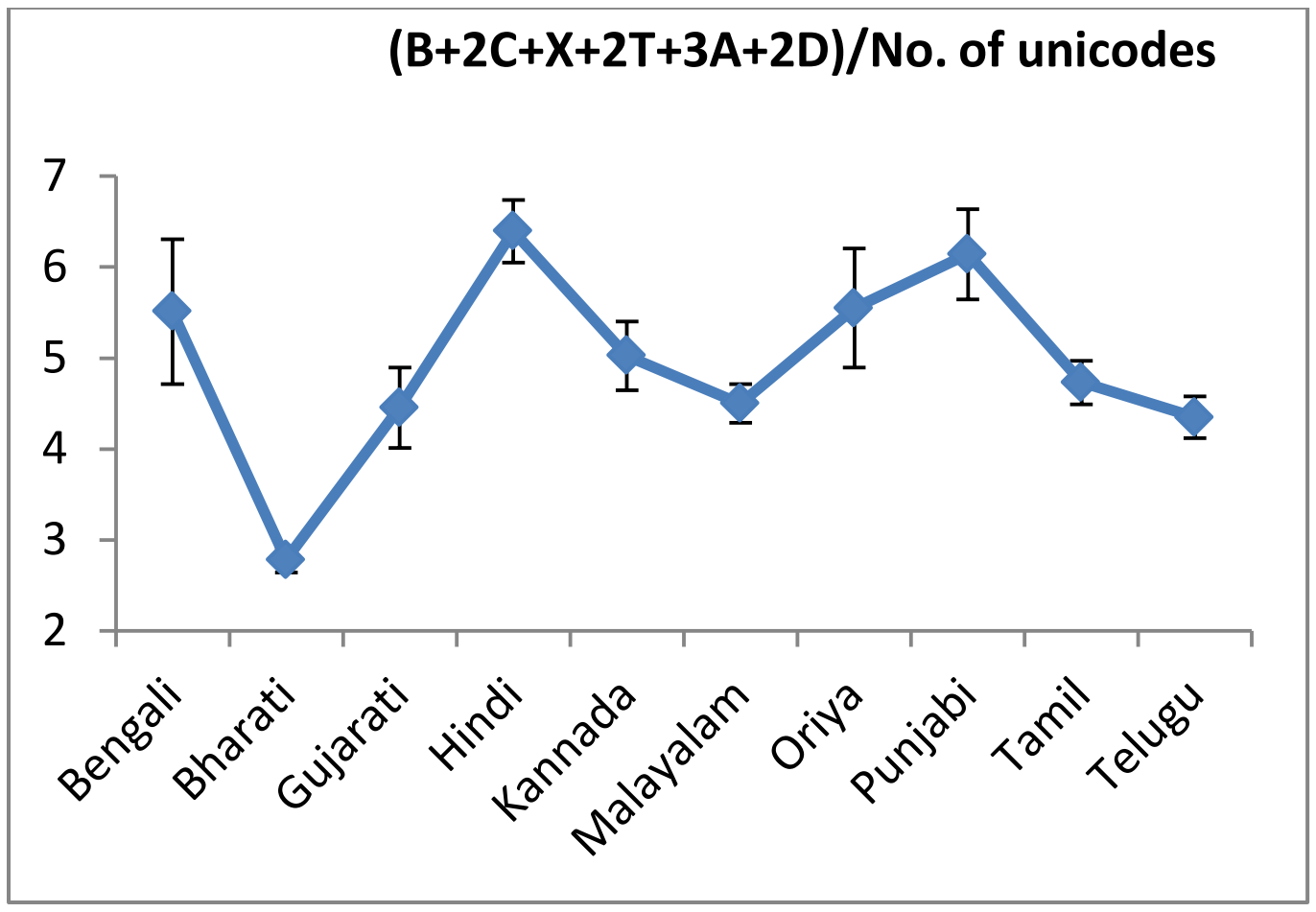

**Figure 15: Complexity #2 is based on only 6 SPs ( B, C, X, T, A, D) excluding E. The least value was observed for Bharati script**

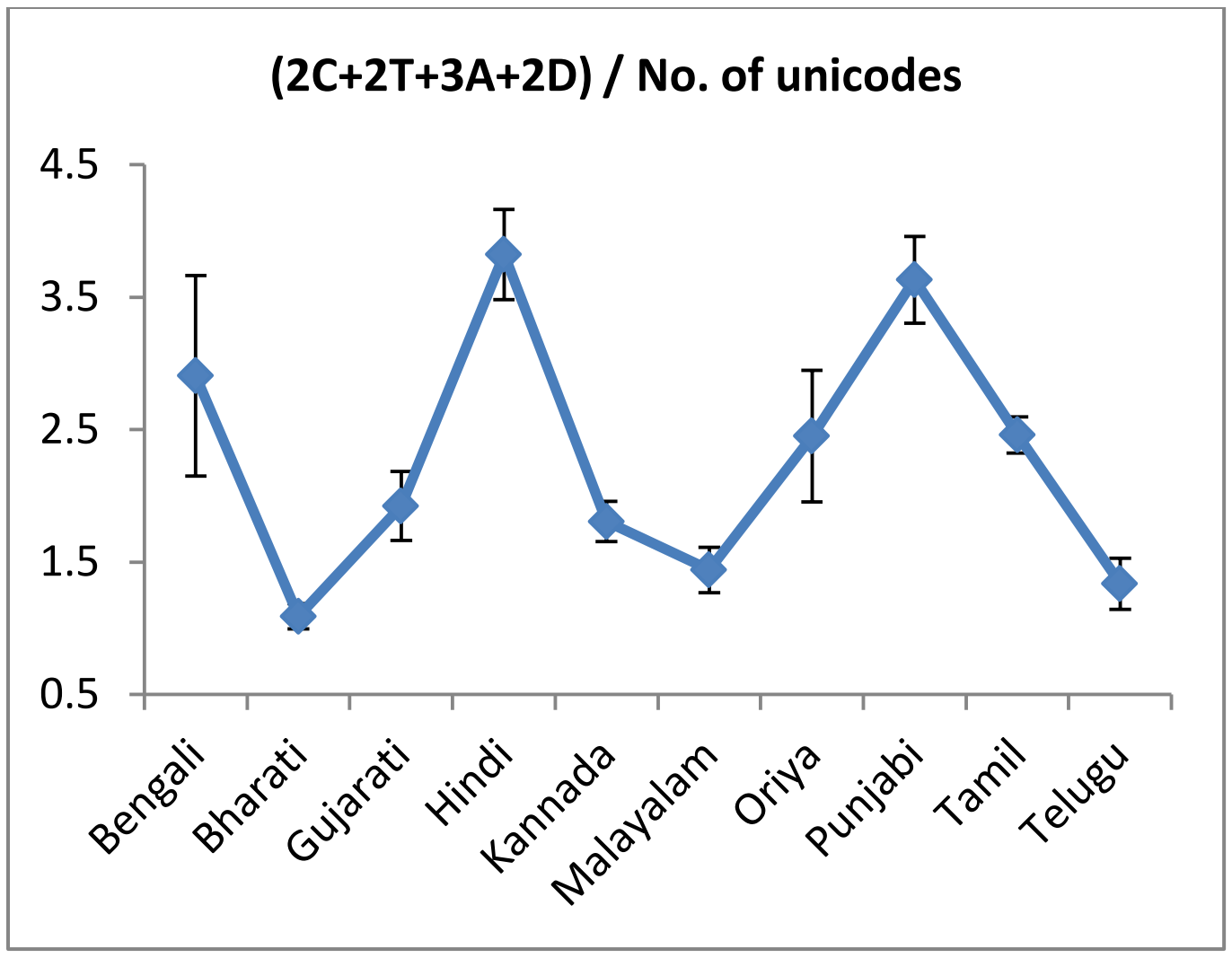

**Figure 16: Complexity #3 (unstable SPs only – C, T, A, D). Bharati script has the lowest value**

## DISCUSSION

Among the major contemporary Indian scripts there are 9 in which the characters are organized as vowels, consonants, Bara Khadi (Consonant-Vowel combinations) etc. Keeping in view this commonness among major Indian scripts, we presented a simple script called Bharati that can represent the 9 major scripts of India. The script derives its simplicity from the underlying principle that guides its design: t*he phonetic organization of Indian language aksharas is strictly reflected in the graphical form of Bharati aksharas.* This being not the case with any of the current scripts, the aksharas of contemporary Indian writing scripts, though often ornate, can pose considerable difficulty to a young learner. Since Bharati aksharas are based on simple organizational rules, the learner can learn the script easily once (s)he understands the phonetic organization.

Another simplifying feature of Bharati aksharas is the fact that the motifs used in the aksharas are drawn from existing Indian scripts or from English. For example, Bharati consonant ‘k’ is constructed by deleting the vertical bar from the Devanagari ‘k’. Bharati consonant ‘cha’ is also constructed similarly. Bharati consonant ‘T’ is identical to ‘T’ (ட) of Tamil script. Other Bharati consonants like ‘n’, ‘m’, ‘v’ and ‘r’ are either the same or slightly altered forms of closest English characters. Thus to anyone who has a knowledge of one of the 9 Indian scripts, and preferably even English, Bharati script offers the comfort of familiarity and therefore facilitates quick learning. Special aksharas that appear in specific scripts (like, for example, ‘zha’ in Tamil or Malayalam, or ‘fa’ in Gurumukhi) are also supported by Bharati. A discussion of special aksharas is omitted here for reasons of space.

One simplifying feature of Bharati character design is the manner in which composite characters are handled. Bharati completely avoids consonant conjuncts, a feature of Indian scripts that leads to creation some of the most complex glyphs. The ideal morphology of composite characters is often a subject of intense debate and the difficulty involved in an easy resolution of this issue serves as a barrier to script reform. Bharati characters include CV-type combinations explicitly. Aksharas of CCV type are broken up as C + halant + C + V. Such handling of consonant conjuncts is adopted by Tamil script, thereby making the script one of the simplest of modern Indian scripts.

Bharati compares favourably with the 9 Indian scripts considered in terms of the complexity measures used in this study. In terms of the number of strokes per Unicode, Bharati figures somewhere in the middle while Tamil and Malayalam take the lowest values (**Error! Reference source not found.**). This is perhaps because Bharati is designed so that component sounds are graphically expressed as segmentable components, which makes the script transparent and lends itself to easy analysis for machine recognition. But the same virtue leads to a script with a greater number of strokes per Unicode. But that disadvantage is offset in other measures of complexity. Bharati has the shortest curvelength per Unicode because the script is designed so that some of the simplest possible glyphs are used to represent any given sound.

Stability Index denotes the relative measure of presence of stable SPs over unstable SPs. Bharati script has the largest Stability Index among the 10 scripts. This is because Bharati glyphs are designed to avoid unstable SPs to the extent possible. For similar reasons Bharati script is found to have lowest values for the three complexity measures considered – Complexity #1 (**Error! Reference source not found.**), Complexity #2 (**Error! Reference source not found.**), Complexity #3 (**Error! Reference source not found.**).

Bharati script opens up the possibility of using a common script across the face of India. Similar to the situation in Europe, a common script across India can eliminate many bottlenecks in communication. It is important to point out that since Bharati is only a script; it does not affect Indian languages in any negative fashion. On the other hand, the growing number of next generation Indians, who can speak a certain Indian language but cannot read or write in the corresponding script, will benefit from adoption of a common script for most Indian languages.

The script derives its simplicity from the underlying principle that guides its design: the phonetic organization of Indian language aksharas is strictly reflected in the graphical form of Bharati aksharas. In any of the current scripts, there is no significant correlation between the shape of characters and the sound of characters. Hence, the aksharas of contemporary Indian writing scripts, though often ornate, can pose considerable difficulty to a young learner. Since Bharati aksharas are based on simple organizational rules, the learner of the script can learn the script easily simply based on the phonetic organization. The complexity of Bharati is proven to be least compared to major Indian scripts considering different measures.

Thus, Bharati script offers the comfort of familiarity and therefore facilitates quick learning. Bharati script opens up the possibility of using a common script across the face of India. Similar to the situation in Europe, a common script across India can eliminate many bottlenecks in communication. It is important to point out that since Bharati is only a script, it does not affect Indian languages in any negative fashion. On the other hand, the growing number of next generation Indians, who can speak a certain Indian language but cannot read or write in the corresponding script, will benefit from adoption of a common script for most Indian languages.

| | | | | |
|---|---|---|---|---|
| **एर्नाकुलम** | এরনাকুলম | એર્નાકુલમ | ಎರ್ನಾಕುಲಂ | എര്ണാകുളം |
| **ऐजोल** | আইবল | ઐજોલ | ಐಜೋಲ್ | ഐജോള് |
| **अंबरनाथ** | আম্বরনাথ | અંબરનાથ | ಅಂಬರ್ನಾಥ್ | അമ്പര്ണത് |
| **झारखंड** | ঝাড়খণ্ড | ઝારખંડ | ಜಾರ್ಖಂಡ್ | ജാർഖണ്ഡ് |
| **श्रीहरीकोटा** | শ্রীহরিকোটা | શ્રીહરિકોટા | ಶ್ರೀಹರಿಕೋಟ | ശ്രിഹാരികോ<br>താ |
| **ओडिशा** | ওড়িশা | ઓડિશા | ಒಡಿಶಾ | ഒഡീഷ |
| छत्तीसगढ | ছত্তিশগড় | છત્તીસગઢ | ಛತ್ತೀಸ್ಗಢ | ഛത്തീസ്ഢ് |
| भरूच | ভারুচ | ભરૂચ | ಭರೂಚ್ | ബറൂച്ച് |
| ठाणे | থানে | થાણે | ಥಾಣೆ | താനെ |
| अमृतसर | অম্রুতসর | અમૃતસર | ಅಮೃತ್ಸರ್ | അമൃതസര് |
| आग्रा | আগ্রা | આગ્રા | ಆಗ್ರಾ | ആഗ്ര |
| इटानगर | ইটানগর | ઇટાનગર | ಇಟಾನಗರ | ഇതനഗര് |
| मुंबई | মুম্বাই | મુંબઇ | ಮುಂಬೈ | മുംബൈ |
| ऊम्बरगाव | অমবড়গাব | ઉંબરગાવ | ಉಬರ್ಗಾಓನ್ | ഉംബര്ഗവ് |
| फतेहपुर<br>सिक्री | ফতেহপুর<br>সিক্রী | ફતેહપુર સિક્રી | ಫತೇಪುರ್<br>ಸಿಕ್ರಿ | ഫത്തേപൂർ<br>സിക്രി |

| Devnagari | Bengali | Gujarati | Kannada | Malayalam |
|---|---|---|---|---|
| **राजस्थान** | রাজস্থান | રાજસ્થાન | ರಾಜಸ್ಥಾನ | രാജസ്ഥാൻ |
| **औरंगाबाद** | ঔরঙ্গাবাদ | ઔરંગાબાદ | ಔರಂಗಾಬಾದ್ | ഔറംഗബാദ് |
| **उदयपूर** | উদয়পুর | ઉદયપુર | ಉದಯ್ಪುರ | ദയ്പൂർ |
| **सिंधुदुर्ग** | সিন্ধুদূর্গ | સિંધુદુર્ગ | ಸಿಂಧುದುರ್ಗ | സിന്ധുദുർഗ് |
| **मेघालय** | মেঘালয় | મેઘાલય | ಮೇಘಾಲಯ | മേഘാലയ |

**Table 3: Names of twenty cities written in nine Indian scripts**

| Devnagari | Oriya | Punjabi | Tamil | Telugu |
|---|---|---|---|---|
| **राजस्थान** | ରାଜସ୍ଥାନ୍ | ਰਾਜਸਥਾਨ | ராஜஸ்தான் | రాజస్థాన్ |
| **औरंगाबाद** | ଔରଙ୍ଗବଦ୍ | ਔਰੰਗਾਬਾਦ | அவுரங்காபாத் | ఔరంగాబాద్ |
| **उदयपूर** | ଉଦୈପୁର୍ | ਉਦੈਪੁਰ | உதய்பூர் | ఉదయపూర్ |
| **सिंधुदुर्ग** | ସିଂଧୁଦୁର୍ଗ୍ | ਸਿੰਦੂਦੁਰ੍ਗ | சிந்துடுர்க் | సింధుదుర్గ్ |
| **मेघालय** | ମେଘଲଯ | ਮੇਘਾਲਯ | மேகாலயா | మేఘాలయ |
| **एर्नाकुलम** | ଏରନକୁଲମ୍ | ਏਰਨਾਕੁਲਮ | எர்ணாகுளம் | ఎర్నాకులం |
| **ऐजोल** | ଇଜୋଲ | ਐਜੋਲ | அய் ஜோ ல | ఏయిసాల్ |
| **अंबरनाथ** | ଅମବର୍ନଥ୍ | ਅਮ੍ਬਰਨਾਥ | அம்பர்னத் | అంబర్నాథ్ |
| **झारखंड** | ଝରଖଣ୍ଡ | ਝਾਰਖੰਡ | ஜார்கண்ட் | జార్ఖండ్ |
| **श्रीहरीकोटा** | ଶ୍ରୀହରିକୋତା | ਸ਼੍ਰੀਹਰੀਕੋਤਾ | ஶ்ரீஹரிகோட்டா | శ్రీహరికోట |
| **ओडिशा** | ଓରିସା | ਓਦੀਸ਼ਾ | ஒடிசா | ఒడిషా |
| छत्तीसगढ | ଛତିସଗଢ | ਛੱਤੀਸਗੜ੍ਹ | சட்டீஸ்கர் | ఛత్తీస్గఢ్ |
| भरूच | ଭରୁଚ | ਭਰੁਚ | பருச் | బారుచ్ |
| ठाणे | ଥାନେ | ਠਾਣੇ | தானே | థానే |
| अमृतसर | ଅମୃତସର୍ | ਅੰਮ੍ਰਿਤਸਰ | அம்ருட்சர் | అమృత్సర్ |
| आग्रा | ଆଗ୍ରା | ਆਗਰਾ | ஆக்ரா | ఆగ్ర |
| इटानगर | ଇତାନଗର୍ | ਇਟਾਨਗਰ | இட்டாநகர் | ఇటానగర్ |
| मुंबई | ମୁମ୍ବଇ | ਮੁੰਬਈ | மும்பை | ముంబై |
| ऊम्बरगाव | ଉମବର୍ଗାବ୍ | ਉਮਬਰਗਾਵ | உம்பர்காவ் | ఉంబరగవ్ |
| फतेहपुर सिक्री | ଫତେହପୁର୍ ସିକରୀ | ਫਤਿਹਪੁਰ ਸੀਕਰੀ | பதேபூர் சிக்ரி | ఫతేపూర్ సిక్రీ |